\documentclass{isprs}
\usepackage[T1]{fontenc}
\usepackage[utf8]{inputenc}
\usepackage{setspace}
\usepackage{geometry} 
\usepackage{epstopdf}
\usepackage{hyperref}
\usepackage{booktabs}
\usepackage{tabularx}
\usepackage[labelsep=period]{caption}  
\usepackage{subcaption}
\usepackage{cleveref}

\newcolumntype{Y}{>{\centering\arraybackslash}X}

\begin{document}

\title{ON THE USABILITY OF DEEP NETWORKS FOR OBJECT-BASED IMAGE ANALYSIS}

\author{%
Nicolas Audebert\textsuperscript{a, b}, Bertrand Le Saux\textsuperscript{a}, Sébastien Lefèvre\textsuperscript{b}
}%

\address
{
	\textsuperscript{a }ONERA, \textit{The French Aerospace Lab}, F-91761 Palaiseau, France (nicolas.audebert,bertrand.le\_saux)@onera.fr\\
	\textsuperscript{b }Univ. Bretagne-Sud, UMR 6074, IRISA, F-56000 Vannes, France - sebastien.lefevre@irisa.fr\\
}


\commission{}{} 
\workinggroup{} 
\icwg{}   

\abstract
{
As computer vision before, remote sensing has been radically changed by the introduction of Convolution Neural Networks. Land cover use, object detection and scene understanding in aerial images rely more and more on deep learning to achieve new state-of-the-art results. Recent architectures such as Fully Convolutional Networks \cite{long_fully_2015} can even produce pixel level annotations for semantic mapping. In this work, we show how to use such deep networks to detect, segment and classify different varieties of wheeled vehicles in aerial images from the ISPRS Potsdam dataset. This allows us to tackle object detection and classification on a complex dataset made up of visually similar classes, and to demonstrate the relevance of such a subclass modeling approach. Especially, we want to show that deep learning is also suitable for object-oriented analysis of Earth Observation data. First, we train a FCN variant on the ISPRS Potsdam dataset and show how the learnt semantic maps can be used to extract precise segmentation of vehicles, which allow us studying the repartition of vehicles in the city. Second, we train a CNN to perform vehicle classification on the VEDAI \cite{razakarivony_vehicle_2016} dataset, and transfer its knowledge to classify candidate segmented vehicles on the Potsdam dataset.
}

\keywords{deep learning, vehicle detection, semantic segmentation, object classification}

\maketitle


\section{INTRODUCTION}\label{sec:introduction}

Deep learning for computer vision grows more popular every year, especially thanks to Convolutional Neural Networks (CNN) that are able to learn powerful and expressive descriptors from images for a large range of tasks: classification, segmentation, detection \ldots This ubiquity of CNN in computer vision is now starting to affect remote sensing as well, as they can tackle many tasks such as land use classification or object detection in aerial images. Moreover, new architectures have appeared, derived from Fully Convolutional Networks \cite{long_fully_2015}, able to output dense pixel-wise annotations and thus able to achieve fine-grained classification. Such architectures have quickly become state-of-the-art for popular datasets such as PASCAL VOC2012 \cite{everingham_pascal_2014} and Microsoft COCO \cite{lin_microsoft_2014}. In an Earth Observation context, these FCN models are now especially appealing, as dense prediction allows us performing semantic mapping without requiring any pre-processing tricks. Therefore, using FCN for Earth Observation means we can shift from superpixel segmentation and region-based classification \cite{lagrange_benchmarking_2015,audebert_how_2016,nogueira_towards_2016} to fully supervised semantic segmentation \cite{marmanis_semantic_2016}.

FCN models have been successfully applied for remote sensing data analysis, notably land cover mapping on urban areas \cite{marmanis_semantic_2016,paisitkriangkrai_effective_2015}. For example, FCN-based models are now the state-of-the-art on the ISPRS Vaihingen Semantic Labeling dataset \cite{rottensteiner_isprs_2012,cramer_dgpf_2010}. Therefore, even though remote sensing images do not share the same structure as natural images, traditional computer vision deep networks are able to successfully extract semantics from them, which was already known for deep CNN-based classifiers \cite{penatti_deep_2015}. This encourages us to investigate further: can we use deep networks to tackle an especially hard remote sensing task, namely object segmentation ? Therefore, this work focuses on using deep convolutional models for segmentation and classification of vehicles using optical remote sensing data.

To tackle this problem, we design a two-step pipeline for segmentation and classification of vehicles in aerial images. First, we use the SegNet architecture \cite{badrinarayanan_segnet:_2015} for semantic segmentation on the ISPRS Potsdam dataset. This allows us generating a pixel-level mask on which we can extract connected components to detect the vehicle instances. Then, using a CNN trained on vehicle classification using the VEDAI dataset \cite{razakarivony_vehicle_2016}, we classify each instance to infer the vehicle type and to eliminate false positives. We then show how to exploit this information to provide new pieces of data about vehicle types and vehicle repartition in the scene.

Our work is closesly related to \cite{marmanis_semantic_2016,paisitkriangkrai_effective_2015} who also use Fully Convolutional Network for urban area mapping. However we focus only on small objects, namely vehicles, although we use the full semantic mapping as an intermediate product in our pipeline. Moreover, regarding \cite{paisitkriangkrai_effective_2015}, our work rely solely on the deeply learnt representation of the optical data, as we do not include any expert features in the process. On the vehicle detection task, \cite{chen_vehicle_2014} tried to detect vehicle in satellite images using deep CNN, but only regressed the bounding boxes. On the contrary, our pipeline is able to infer the precise object segmentation, which could be regressed into a bounding box if needed. In addition, we also classify the vehicles into several subcategories using a classifier trained on a more generic dataset, thus performing transfer learning.

\section{PROPOSED METHOD}\label{sec:method}

\begin{figure*}
  \includegraphics[width=\textwidth]{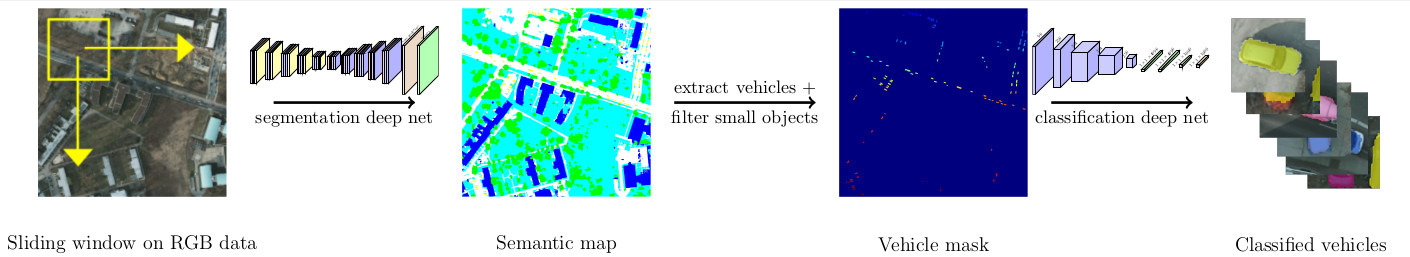}
  \caption{Illustration of our vehicle segmentation + classification pipeline}
  \label{fig:vehicle_extraction}
\end{figure*}

\subsection{Semantic segmentation}\label{sec:segnet}
Many deep network architectures are available for semantic segmentation, including the original FCN \cite{long_fully_2015} or many variants such as DeepLab \cite{liang-chieh_semantic_2015}. We choose to use the SegNet architecture \cite{badrinarayanan_segnet:_2015}, since it is well-balanced between accuracy and computational cost. SegNet's symmetrical structure and its use of pooling/unpooling layers is very effective for precise relocalisation of features according to its authors. Indeed, preliminary tests underlined that SegNet performs very well, including for small object localization. Results with other architectures such as FCN reported either no improvement or an unsignificant improvement. However, note that any deep network trained for semantic segmentation can be used in our pipeline.

SegNet is based on the convolutional layers of the VGG-16 model \cite{simonyan_very_2014}. VGG-16 was designed for the ILSRVC competition and trained on the ImageNet dataset \cite{russakovsky_imagenet_2015}. Each block of the encoder is comprised of 2 or 3 convolutional layers followed by batch normalization \cite{ioffe_batch_2015} and rectified linear units. A maxpooling layer reduces the dimensions between each block. In the decoder, symetrical operations are applied in the reverse order. The unpooling operation replaces the pooling in the decoder: it relocates the value of the activations into the mask of the maximum values (``\textit{argmax}'') computed at the pooling stage. Such an upsampling results in a sparse activation map that is densified by the consecutive decoding convolutions. This allows us upsampling the feature activations up to the original input size, so that after the decoder, the final output feature maps have the same dimensions as the input, which is necessary for pixel-level inference.

We initialize the weights of the encoder using VGG-16's weights trained on ImageNet. By doing so, we leverage the expressiveness of the filters learned by VGG-16 when training on this very diverse dataset. This initialization allows us improving the final segmentation accuracy compared to a random initialization and also makes the model converge faster.

We train SegNet on semantic segmentation on the ISPRS Potsdam dataset. This translates to pixel-wise classification, i.e each pixel is classified as belonging to one class from the ISPRS benchmark: ``impervious surface'', ``building'', ``tree'', ``low vegetation'', ``car'' or ``clutter''. From this semantic map, we can then extract the vehicle mask which labels all the pixels inferred to belong to the ``car'' class.

\subsection{CNN classification}\label{sec:cnn}

After generating the full semantic map from one tile, we can extract the vehicle mask. We perform connected components extraction to find all candidate vehicles in the predicted map. To separate vehicles that are very close to each other in the RGB image and that might belong to the same connected component, we use a morphological opening to open the gap and separate the objects. In addition, to smooth the predictions and eliminate small artifacts and imperfections from SegNet's predicted map, we removed connected components that are too small  to be a vehicle (surface $<$ 32~px). For each vehicle (any remaining connected component), we extract a rectangular patch around its bounding box, including 16~px of spatial context on all sides. This patch is then fed to a CNN classifier trained to infer the vehicle's precise class. The full pipeline is illustrated in \Cref{fig:vehicle_extraction}.

We compare 3 CNN for vehicle classification: LeNet \cite{lecun_gradient-based_1998}, AlexNet \cite{krizhevsky_imagenet_2012} and VGG-16 \cite{simonyan_very_2014}. We use the VEDAI dataset to train these models, so that the Potsdam vehicles will be entirely unseen data. This is to show how knowledge learnt by deep networks on remote sensing data can be transferred efficiently from one dataset to another, under the assumption that provided resolutions (12.5cm) and data sources (RGB) are the same. The classifiers learn to discriminate between the following 11 classes of vehicles from VEDAI, with a very high variability: ``car'', ``camping car'', ``tractor'', ``truck'', ``bike'', ``van'', ``bus'', ``ship'', ``plane'', ``pick up'' and ``other vehicles''.

LeNet is a model introduced by \cite{lecun_gradient-based_1998} designed for character recognition in greyscale image. Our version takes in input a $224\times224$ color image. It is a comparatively small model with few parameters ($\simeq$ 461K parameters) that can be trained very quickly.

AlexNet is a very popular model introduced in \cite{krizhevsky_imagenet_2012} that won the ILSVRC challenge in 2012. It takes in input a $227\times227$ color image. As there are three final fully connected layers, AlexNet is a relatively big network with 61M parameters. We fine-tune the last layer of the reference ImageNet trained implementation of AlexNet.

Finally, VGG-16 is another popular model designed by \cite{simonyan_very_2014}, on which is based the SegNet segmentation network introduced in \Cref{sec:segnet}. It outperformed AlexNet on the ImageNet benchmark in 2014. It takes in input a $224\times224$ color image. VGG-16 is the biggest of our three models with 138M parameters. Once again, we fine-tune only the last layer of the reference weights of VGG-16 trained on ImageNet. It is the biggest and slowest of the three models tested in this work.

\section{EXPERIMENTS}\label{sec:experiments}
\begin{table*}[!htbp]
  \caption{Average CNN accuracies on VEDAI (3 train/test splits)}
  \label{table:classif_vedai}
  \begin{tabularx}{\textwidth}{Y c c c c c c c c c c}
  \toprule
  Model/Class & Car & Truck & Ship & Tractor & Camping car & Van & Pick up & Plane & Other & Global\\
  \midrule
  LeNet & 74.3\% & 54.4\% & 31.0\% & 61.1\% & 85.9\% & 38.3\% & 67.7\% & 13.0\% & 47.5\% & 66.3 $\pm$ 1.7\%\\
  AlexNet & \textbf{91.0\%} & 84.8\% & 81.4\% & 83.3\% & 98.0\% & \textbf{71.1\%} & 85.2\% & 91.4\% & \textbf{77.8\%} & 87.5 $\pm$ 1.5\%\\
  VGG-16 & 90.2\% & \textbf{86.9\%} & \textbf{86.9\%} & \textbf{86.5\%} & \textbf{99.6\%} & \textbf{71.1\%} & \textbf{91.4\%} & \textbf{100.0\%} & 77.2\% & \textbf{89.7} $\pm$ 1.5\%\\
  \bottomrule
  \end{tabularx}
\end{table*}

\subsection{Experimental setup}\label{sec:datasets}
\subsubsection{VEDAI}
This VEDAI dataset \cite{razakarivony_vehicle_2016} is comprised of 1268 RGB tiles ($1024\times1024$px) and the associated IR image at 12.5cm spatial resolution. For each tile, annotations are provided detailing the position of the vehicles and the associated bounding box (defined by its 4~corners). We cross-validated the results on three of the suggested splits of the dataset for the training and testing sets. The training set for the CNN is built by extracting square patches around each vehicle bounding box, padded with 16~px of spatial context on each side. We use data augmentation to increase the number of samples by including rotated ($\frac{\pi}{2} $, $\pi$ and $\frac{3\pi}{2}$) and mirrored versions of the patches containing vehicles.

We train each model during 50 epochs ($\simeq$~1000000~iterations) with a batch size of respectively 128 for LeNet and AlexNet and 32 for VGG-16, and a learning rate of 0.001, divided by 10 after 30 epochs. Training takes about 25 minutes for LeNet, 60 minutes for AlexNet and 10 hours for VGG-16 on NVIDIA K20c GPU.

\subsubsection{ISPRS Potsdam}
This ISPRS Potsdam Semantic Labeling dataset \cite{rottensteiner_isprs_2012} is comprised of 38 RGB tiles ($6000\times6000$~px) and the associated IR and DSM images at 5cm spatial resolution. A comprehensive ground truth is provided for 24 tiles, which are the tiles we will work with. We split randomly the dataset in train/validation/test with 70/10/20 proportions. For a fair comparison of the two datasets (ISPRS Potsdam and VEDAI) in the same framework, we only use the RGB data and downsample the resolution from 5~cm/pixel to 12.5~cm/pixel. However, note that SegNet would perform even better for small object segmentation on the high resolution tile. 

We build our segmentation training set by sliding a window of $128\times128$~px over each high resolution tile, including an overlap of 75\% (32~px stride). For this experiment, we use all the classes from the ground truth. This means that we not only train the model to predict the vehicle mask, but also to assign a label to each pixel according to the ISPRS classes, except ``clutter'': ``impervious surface'', ``building'', ``tree'', ``low vegetation'' and ``car''. The model is trained by Stochastic Gradient Descent for 10 epochs ($\simeq$~100000~iterations) with a batch size of 10 and a learning rate of 0.1, divided by 10 after 3 and 8 epochs. This takes around 24 hours with a NVIDIA K20c GPU.

At testing time, we process the tiles by sliding a window of $128\times128$~px with an overlap of 50\%. For overlapping pixels, we average the multiple predictions on each pixel. Processing one tile takes around 70 seconds on a NVIDIA K20c GPU. The predicted semantic map contains all classes from the ISPRS benchmark. We extract only the mask for the ``car'' class as we are mainly interested in how SegNet behaves for small objects, especially vehicles. We report results before and after removing objects smaller than a predefined threshold.

To compute vehicle classification metrics, we build manually an enhanced ground truth by subdividing the ``car'' class into several subcategories: ``cars'', ``vans'', ``trucks'', and ``pick ups''. We discard other vehicles that are present in the RGB data (especially construction vehicles) but not in the original ground truth.

\subsection{Results}\label{sec:results}
\begin{figure}
  \begin{subfigure}{0.15\textwidth}
    \includegraphics[width=\textwidth]{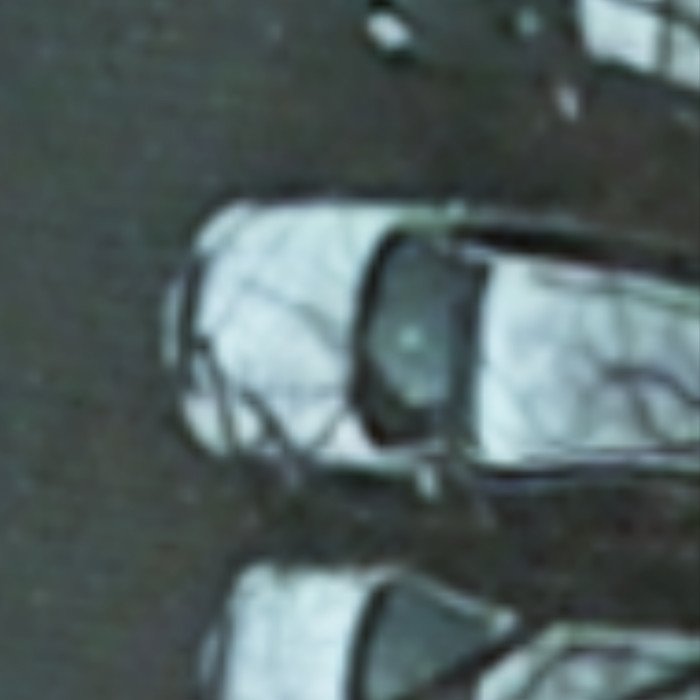}
    \caption{RGB data}
  \end{subfigure}
  \begin{subfigure}{0.15\textwidth}
    \includegraphics[width=\textwidth]{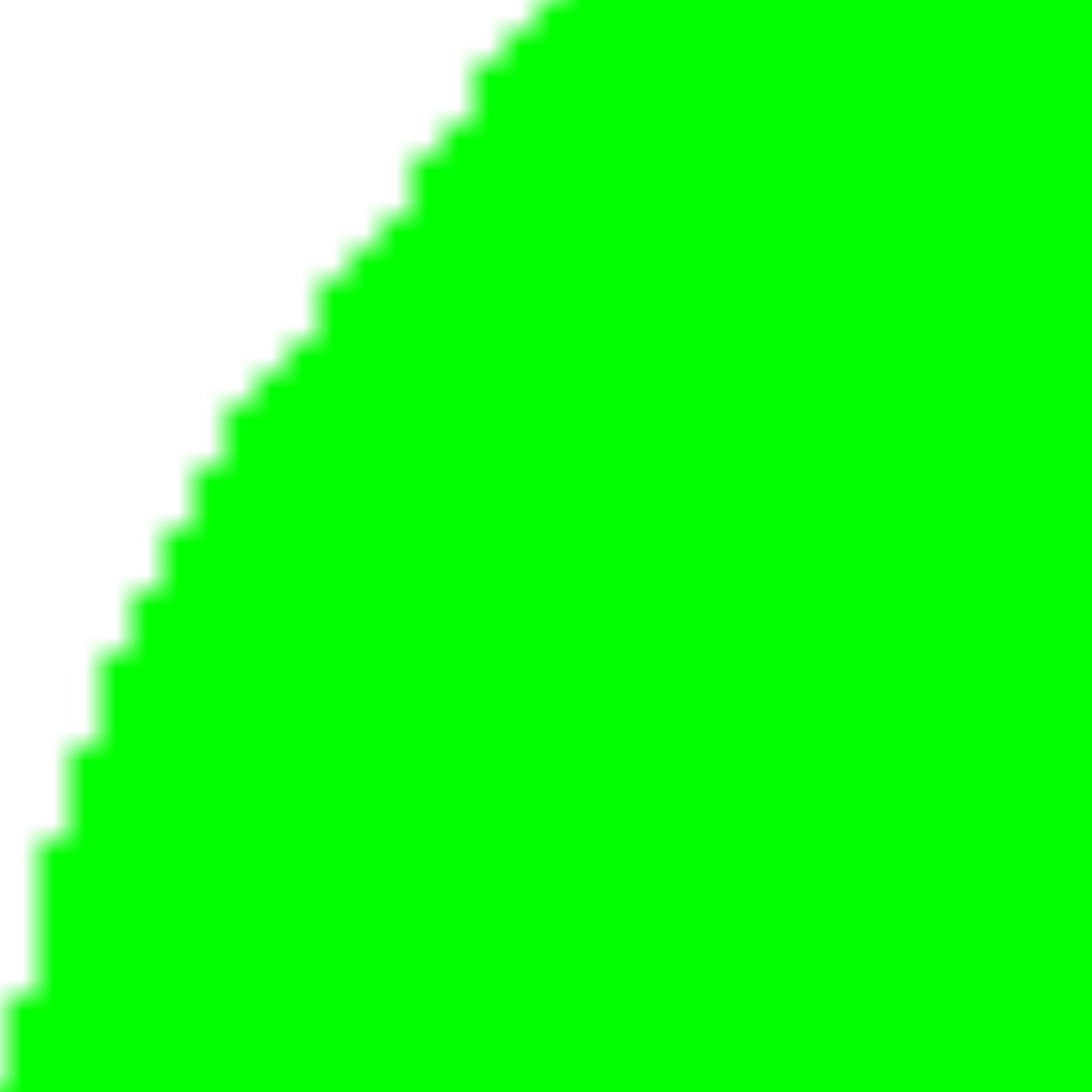}
    \caption{Ground truth}
  \end{subfigure}
  \begin{subfigure}{0.15\textwidth}
    \includegraphics[width=\textwidth]{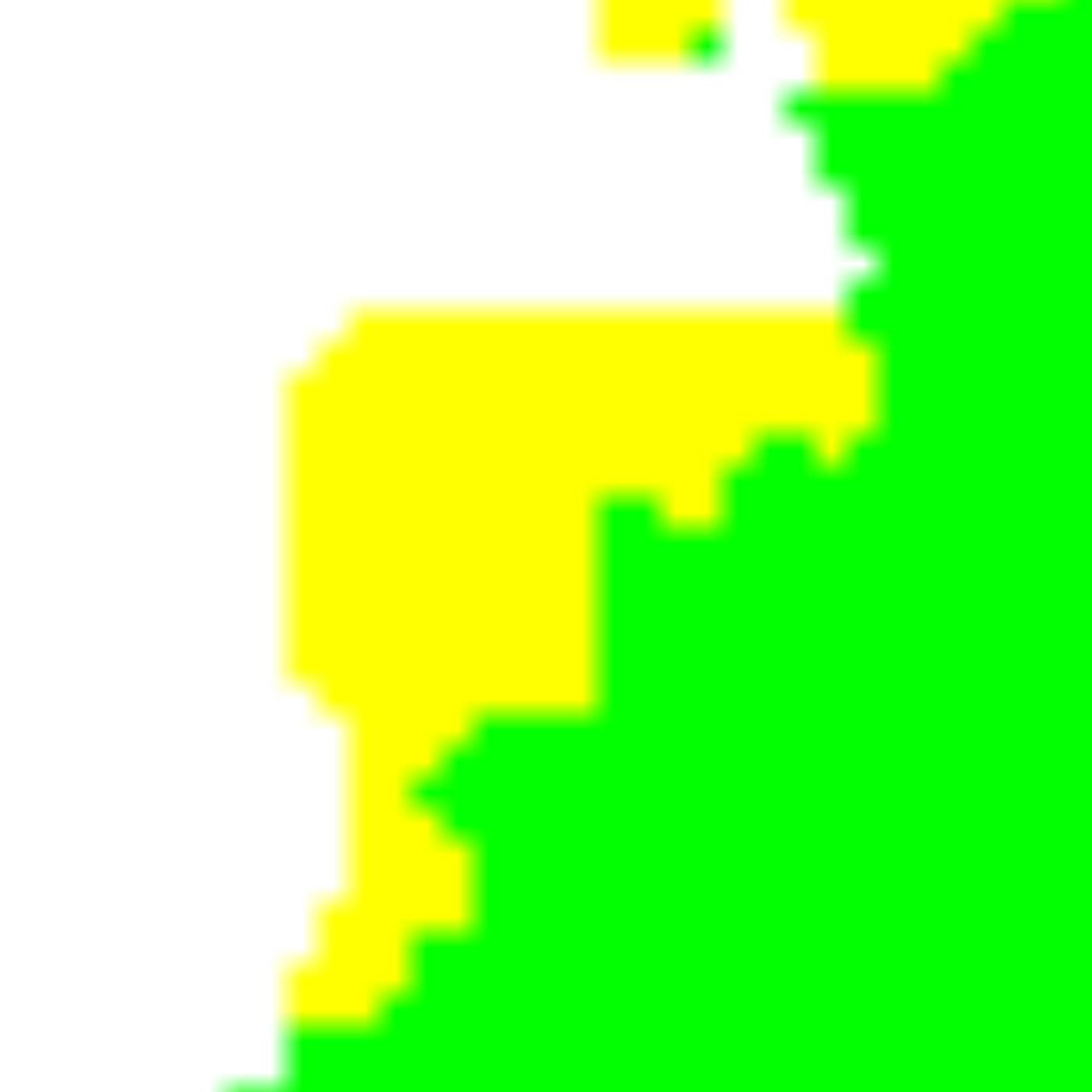}
    \caption{SegNet output}
  \end{subfigure}
  \caption{Car partially occluded by a tree on the Potsdam dataset and misclassification according to the ground truth}
  \label{fig:occlusion}
\end{figure}
\subsubsection{VEDAI}

As detailed in \Cref{table:classif_vedai}, both VGG-16 and AlexNet are able to discriminate between the different classes of the VEDAI dataset, with respectively 89.7\% and 87.5\% global accuracy. LeNet is lagging behind and performs significantly worse than its deeper counterpart, with an accuracy of only 66.3\%. This result is not surprising, as LeNet is a very small network in comparison to both AlexNet and VGG-16 and therefore has a dramatically lower expressive power. The LeNet model performs adequately on ``easy'' classes, such as cars and camping cars, that have low intra-class variability and high inter-class variability. However, it does not work well on vans, which are difficult to distinguish from cars from bird's view. Moreover, classes with high intra-class variability (especially trucks and planes) are often misclassified by LeNet.

Indeed, we argue that as VGG-16 and AlexNet were pre-trained on ImageNet and fine-tuned on VEDAI, these models were already initialized with very powerful and expressive visual descriptors learnt on a highly diverse image dataset. Therefore, both networks benefited from a pre-training phase on a comprehensive dataset of natural images requiring complex filters able to analyze shapes, textures and colors, which helped training on VEDAI, even though ImageNet does not contain any remote sensing image. This fine-grained recognition based on textures and shape allow the CNN to discrminate between classes with a low inter-class variability such as cars and vans, and perform well on classes with high intra-class variability such as planes or trucks. This supports evidence that transfer learning and fine-tuning of ImageNet-trained deep convolutional neural networks is an effective way to process aerial images as shown in previous works \cite{penatti_deep_2015}.

\subsubsection{ISPRS Potsdam}

\paragraph{Vehicle segmentation}

\begin{figure}
  \begin{subfigure}{0.15\textwidth}
    \includegraphics[width=\textwidth]{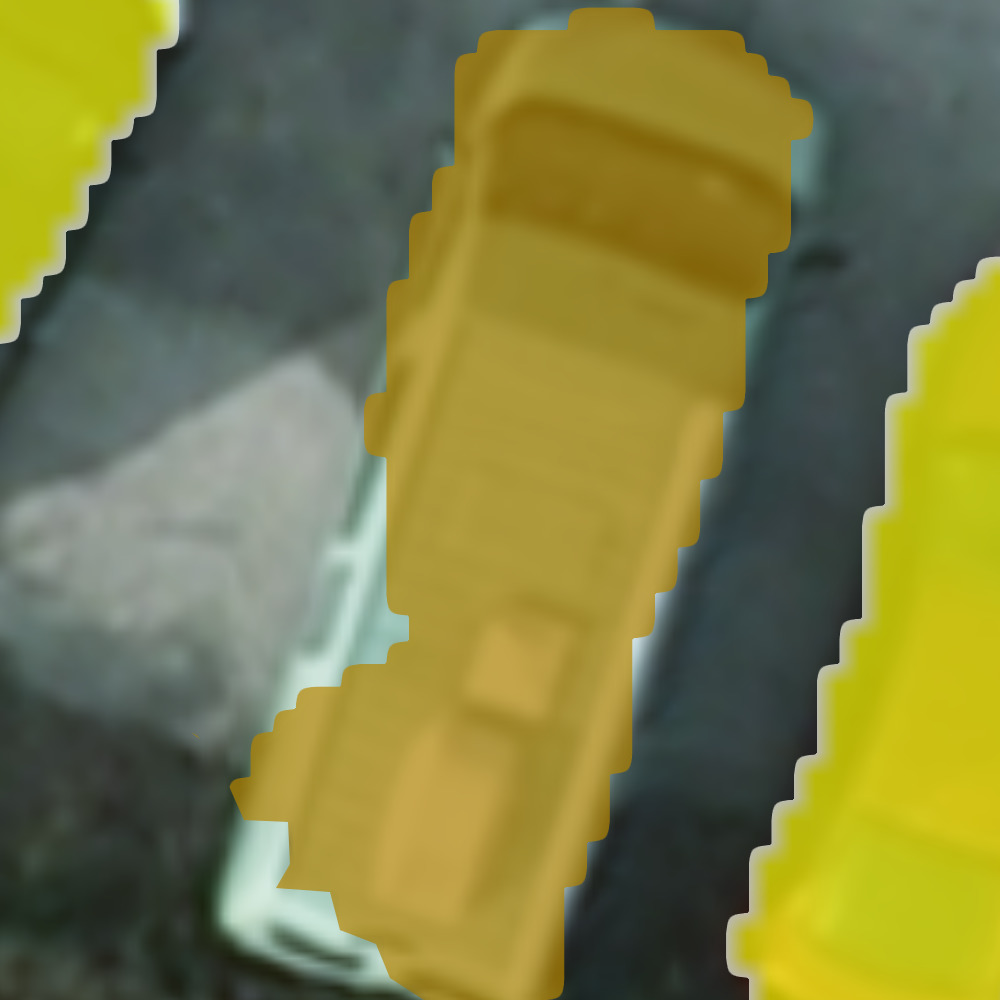}
    \caption{Van misclassified as truck}
  \end{subfigure}
  \begin{subfigure}{0.15\textwidth}
    \includegraphics[width=\textwidth]{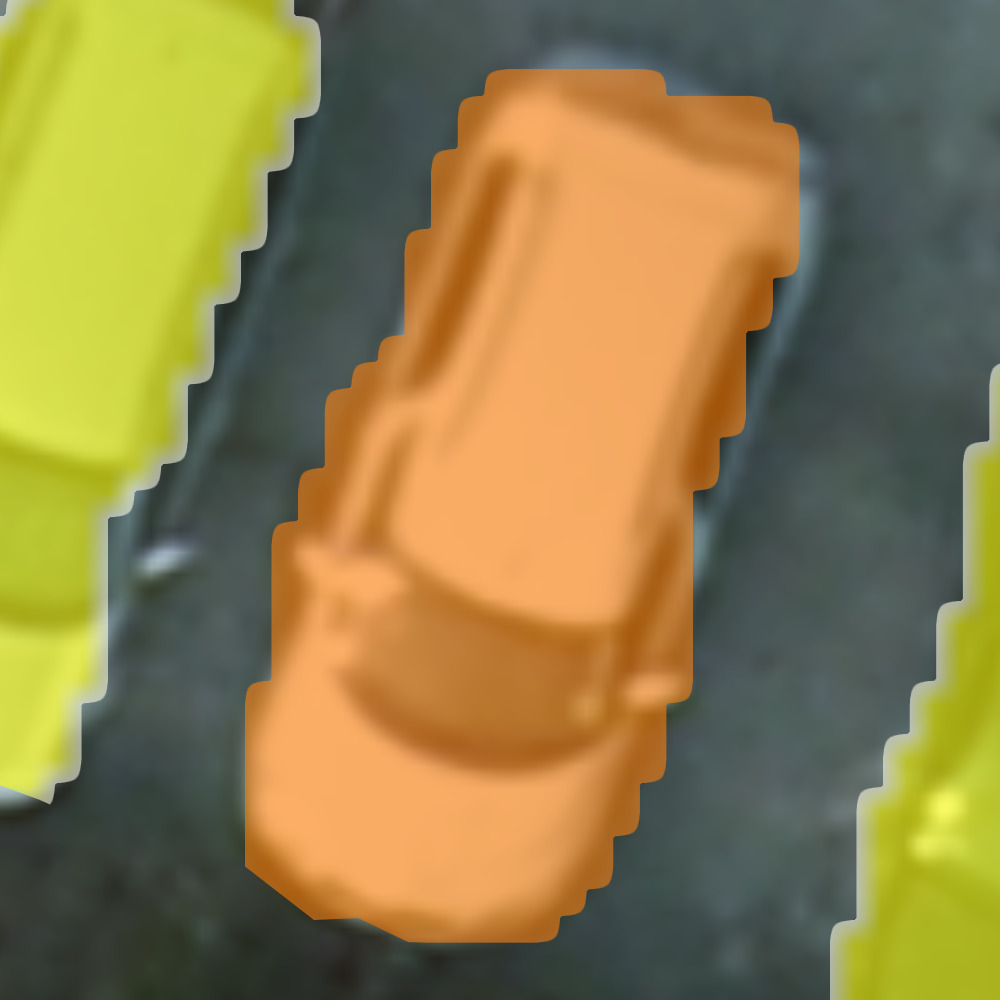}
    \caption{Car misclassified as van}
  \end{subfigure}
  \begin{subfigure}{0.16\textwidth}
    \includegraphics[width=\textwidth]{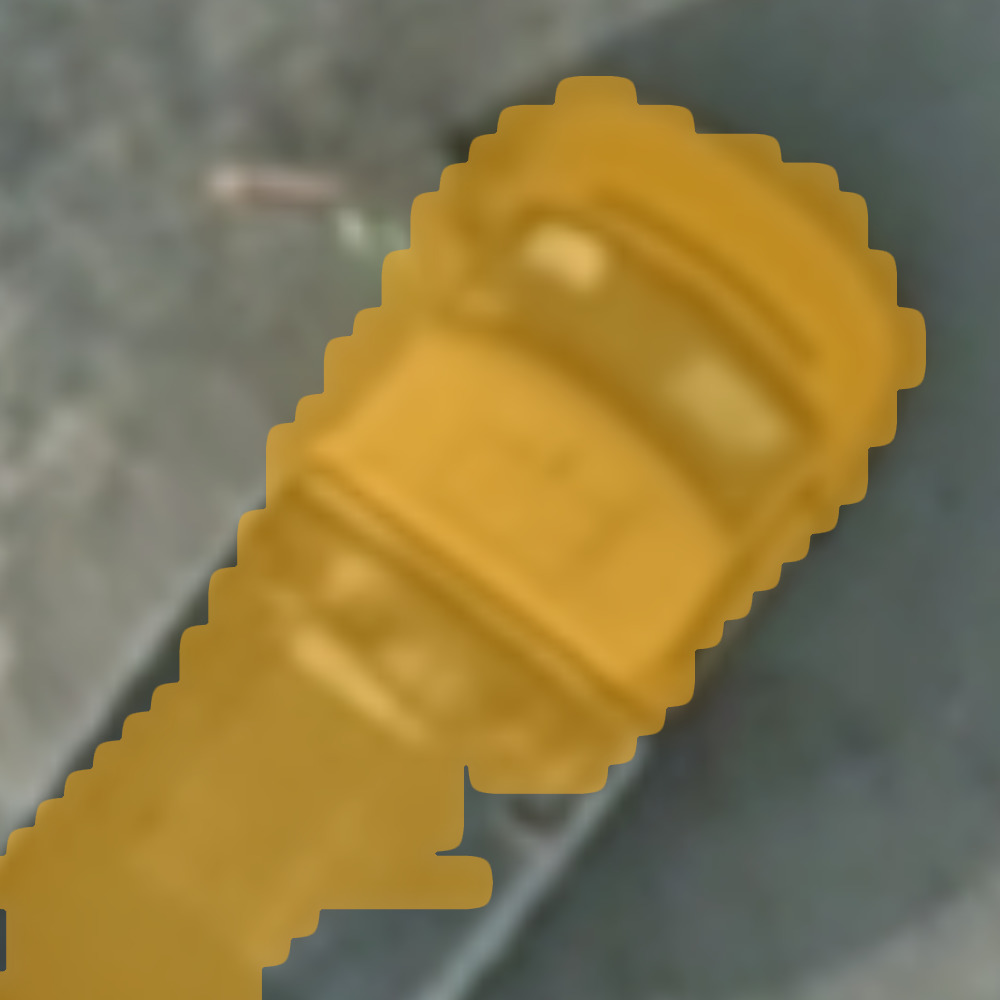}
    \caption{Pick up misclassified as van}
  \end{subfigure}
  \caption{Successful segmentation but misclassified vehicles in the Potsdam dataset}
  \label{fig:negative_examples}
\end{figure}

\begin{figure*}[!t]
  \begin{subfigure}{0.16\textwidth}
    \includegraphics[width=\textwidth]{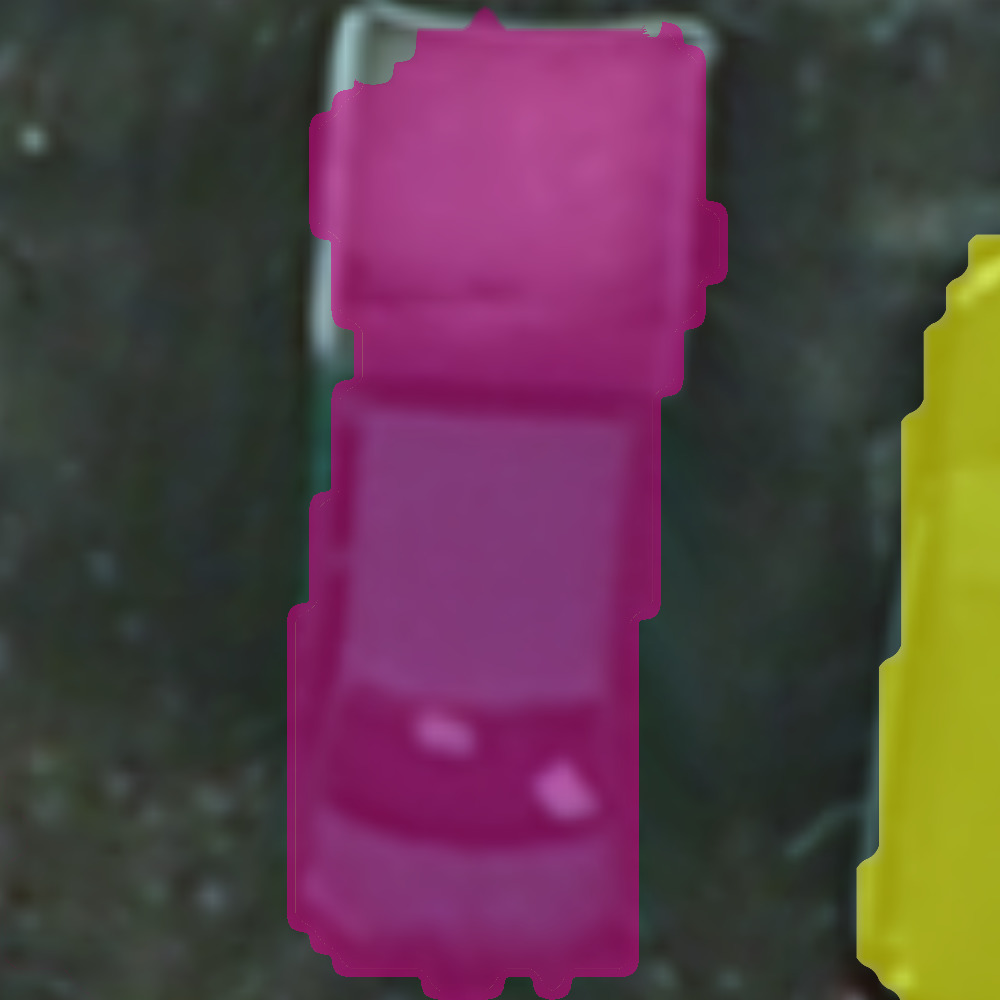}
    \caption{Pick up}
  \end{subfigure}
  \begin{subfigure}{0.16\textwidth}
    \includegraphics[width=\textwidth]{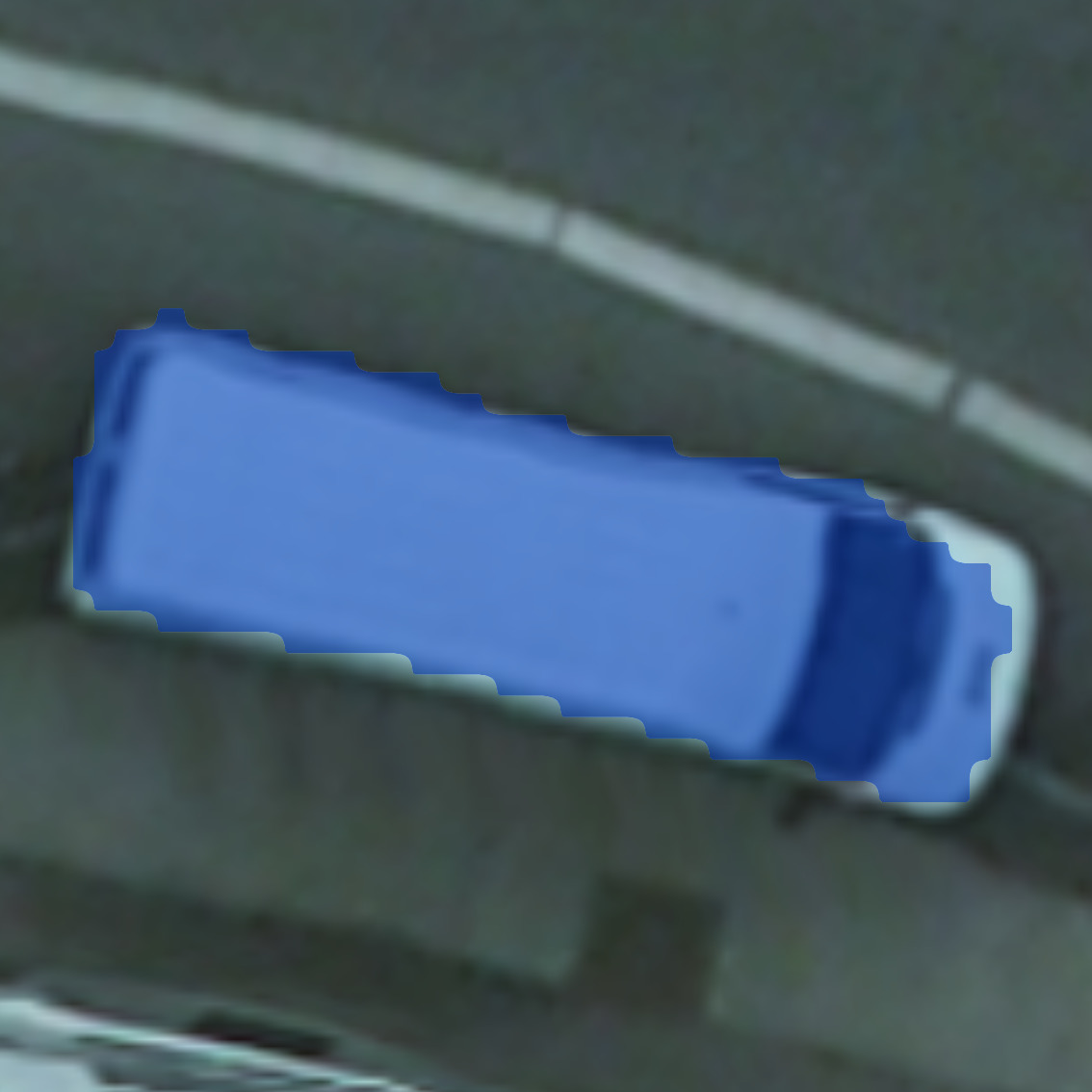}
    \caption{Van}
  \end{subfigure}
  \begin{subfigure}{0.16\textwidth}
    \includegraphics[width=\textwidth]{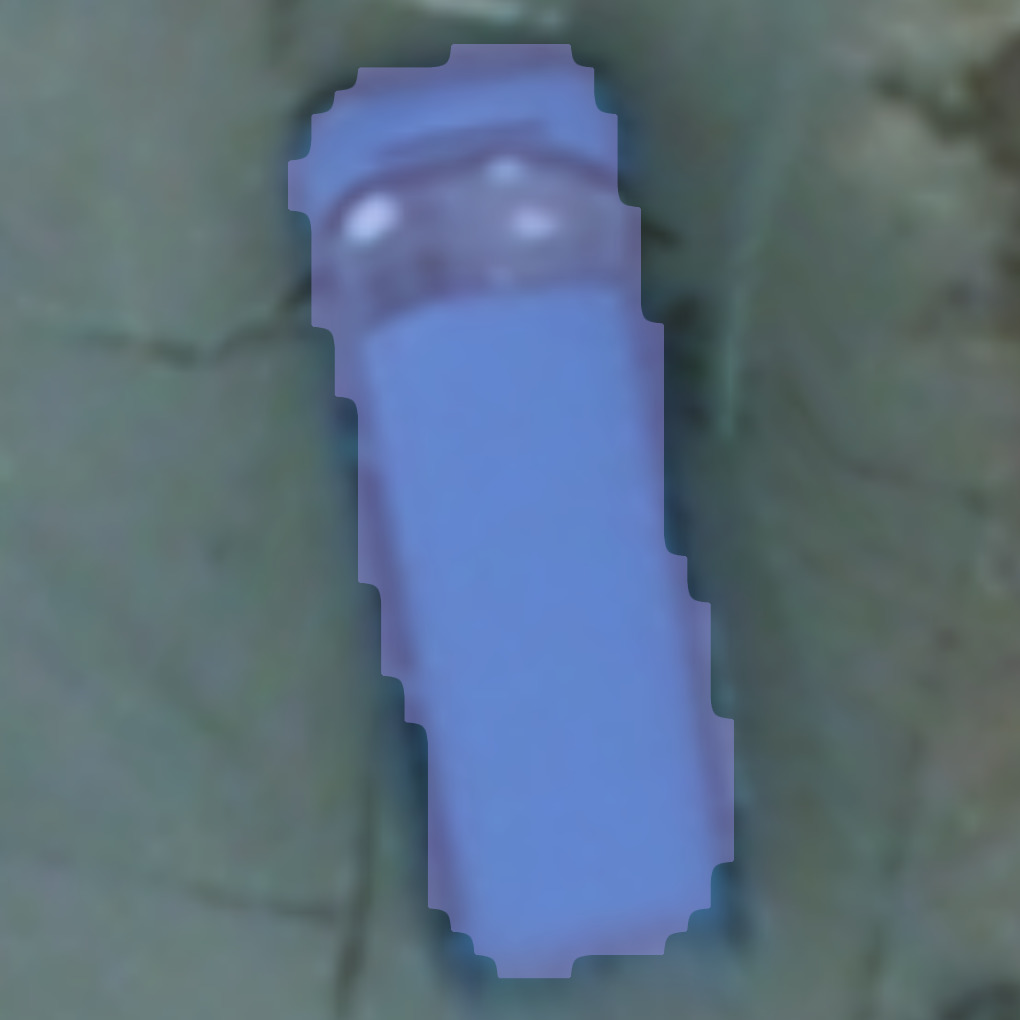}
    \caption{Van}
  \end{subfigure}
  \begin{subfigure}{0.16\textwidth}
    \includegraphics[width=\textwidth]{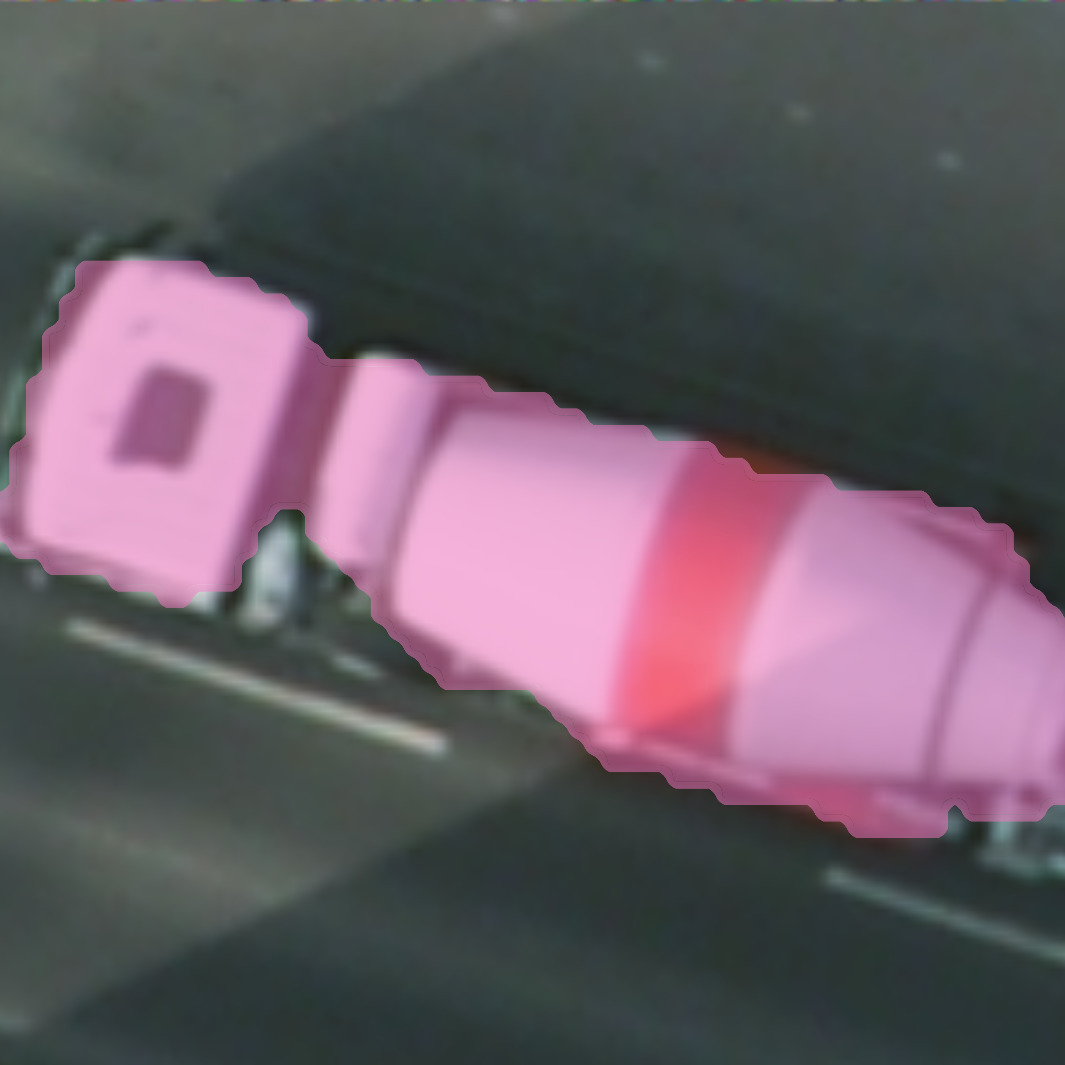}
    \caption{Truck}
  \end{subfigure}
  \begin{subfigure}{0.16\textwidth}
    \includegraphics[width=\textwidth]{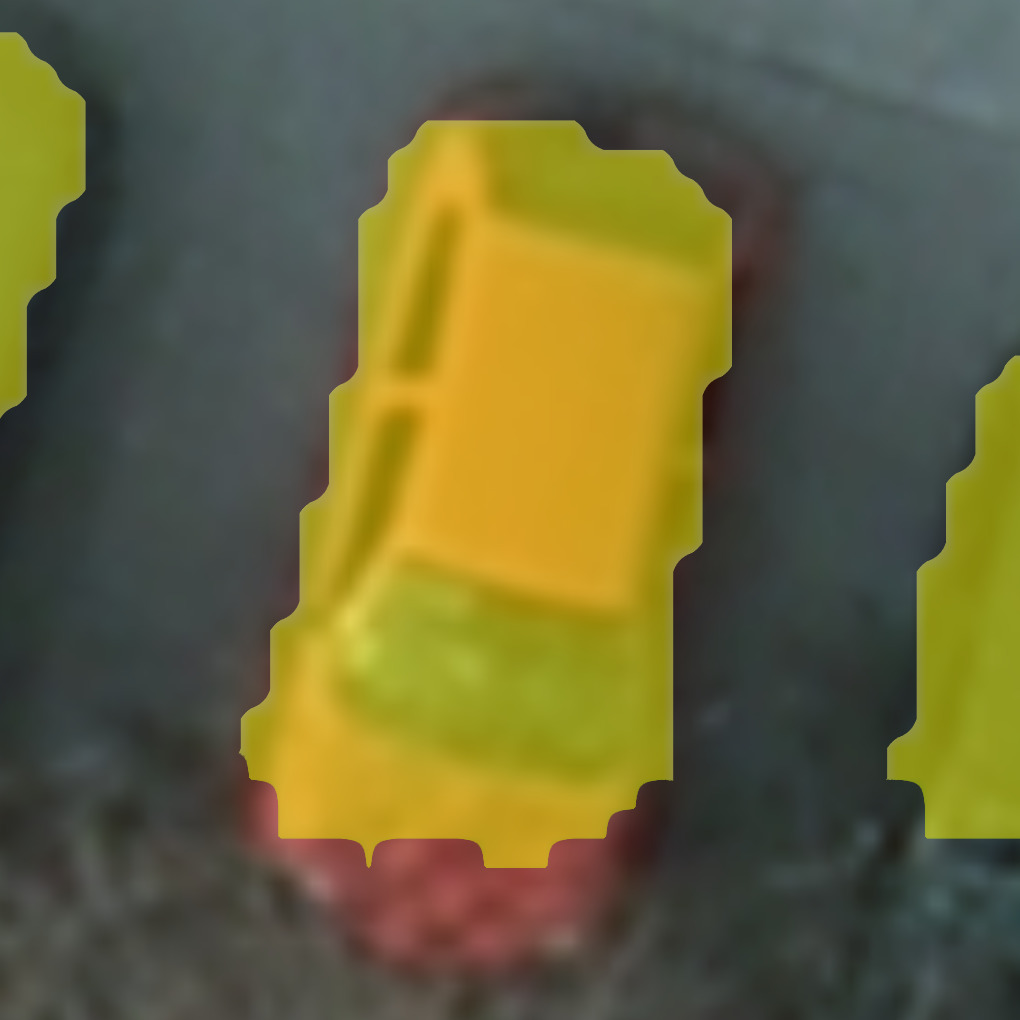}
    \caption{Car}
  \end{subfigure}
  \begin{subfigure}{0.16\textwidth}
    \includegraphics[width=\textwidth]{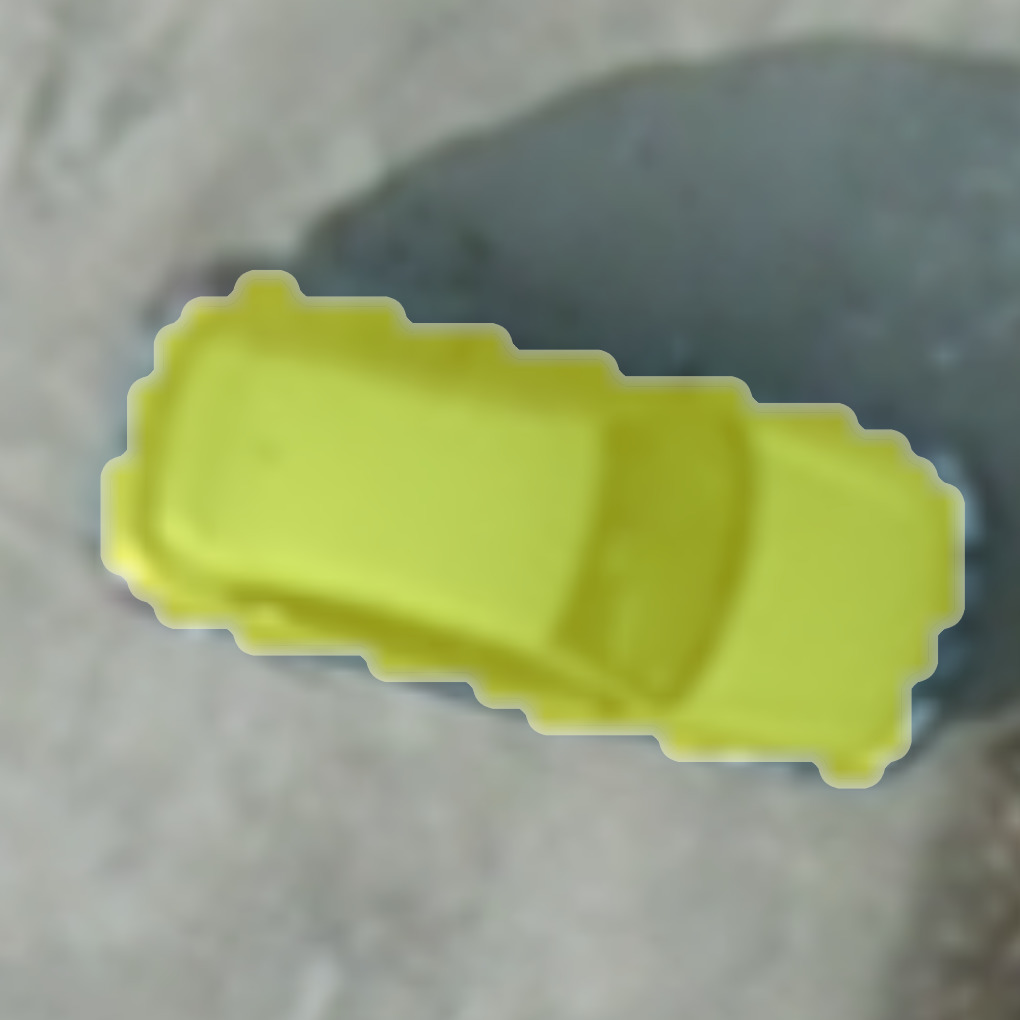}
    \caption{Car}
  \end{subfigure}
  \caption{Successful segmentation and classification of vehicles in the Potsdam dataset}
  \label{fig:positive_examples}
\end{figure*}

\begin{figure*}
  \begin{subfigure}{0.32\textwidth}
    \includegraphics[width=\textwidth]{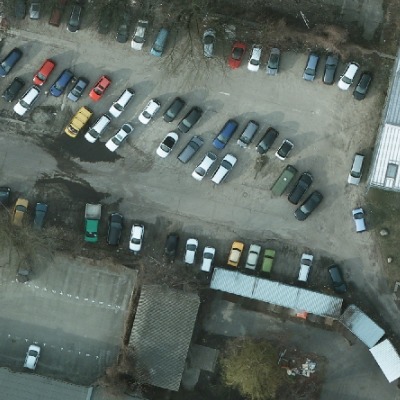}
    \caption{RGB data}
  \end{subfigure}
  \begin{subfigure}{0.32\textwidth}
    \includegraphics[width=\textwidth]{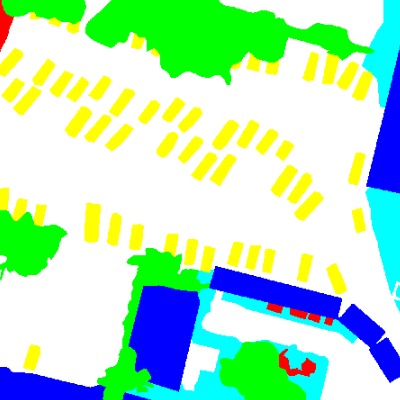}
    \caption{Ground truth}
  \end{subfigure}
  \begin{subfigure}{0.32\textwidth}
    \includegraphics[width=\textwidth]{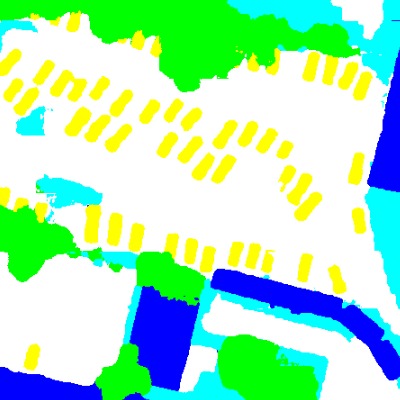}
    \caption{SegNet prediction}
  \end{subfigure}
  \caption{SegNet results on a parking lot (extracted from the ISPRS dataset, best viewed in color)}
  \label{fig:parking_zoom}
\end{figure*}

The overall accuracy of the SegNet semantic mapping on the downsampled ISPRS dataset is 89.5\%, excluding clutter and eroding borders by 3~px. Compared to the state-of-the-art result of 90.3\%\footnote{\url{http://www2.isprs.org/potsdam-2d-semantic-labeling.html}}, this result is quite good considering that we use a downscaled version of the data\footnote{However, we only report our result on a validation set and not on the full testing set, as full semantic mapping of the ISPRS Potsdam dataset is out of the scope of this work.}. When considering only the ``car'' class (which comprises all kinds of terrestrial motorized vehicles), the F1 score at pixel level is 88.4\% (95.5\% with the eroded borders). We also report a pixel-wise precision of 87.8\% and a recall of 89.0\%. This indicates that the vehicle mask generated from SegNet's semantic map should be very accurate and can be used for further object-based analysis. This is supported by qualitative inspection of the semantic maps on car-heavy areas, such as the one presented in \Cref{fig:parking_zoom}.

The mean Intersection over Union (IoU) reaches 75.6\%. However, there is a high variance in the IoU depending on the specific surroundings of each car. Indeed, because of occlusions due to trees, the ground truth sometimes is labelled as tree even if the car is visible in the RGB image. This causes SegNet to misclassify (according to the ground truth) most of the pixels, resulting in a low IoU despite the visual quality of the segmentation. This phenomenon is illustrated by \Cref{fig:occlusion}. However, as SegNet manages to still infer the car's contours, this should not impede too severely later vehicle extraction and classification.

\paragraph{Counting cars}

\begin{table}
  \caption{Vehicle counts in testing tiles of ISPRS Potsdam}
  \label{table:counting}
  \begin{tabularx}{0.5\textwidth}{c Y Y Y Y Y}
  \toprule
  Tile \# & 2\_11 & 7\_12 & 3\_11 & 5\_12 & 7\_10\\
  \midrule
  Cars in ground truth & 110 & 351 & 168 & 428 & 253\\
  Predicted cars & 115 & 342 & 182 & 435 & 257\\
  \bottomrule
  \end{tabularx}
\end{table}

We extract the vehicle mask from the semantic map of Potsdam generated by SegNet. Each connected component is expected to be a vehicle instance. We count the number of predicted instances in each tile of the testing set. Detailed results are reported in \Cref{table:counting} and show that our model can be used to estimate precisely the number of vehicles in the original tile with an average relative error of less than 4\%. 

\begin{figure*}
  \centering
  \begin{subfigure}{0.35\textwidth}
    \includegraphics[width=\textwidth]{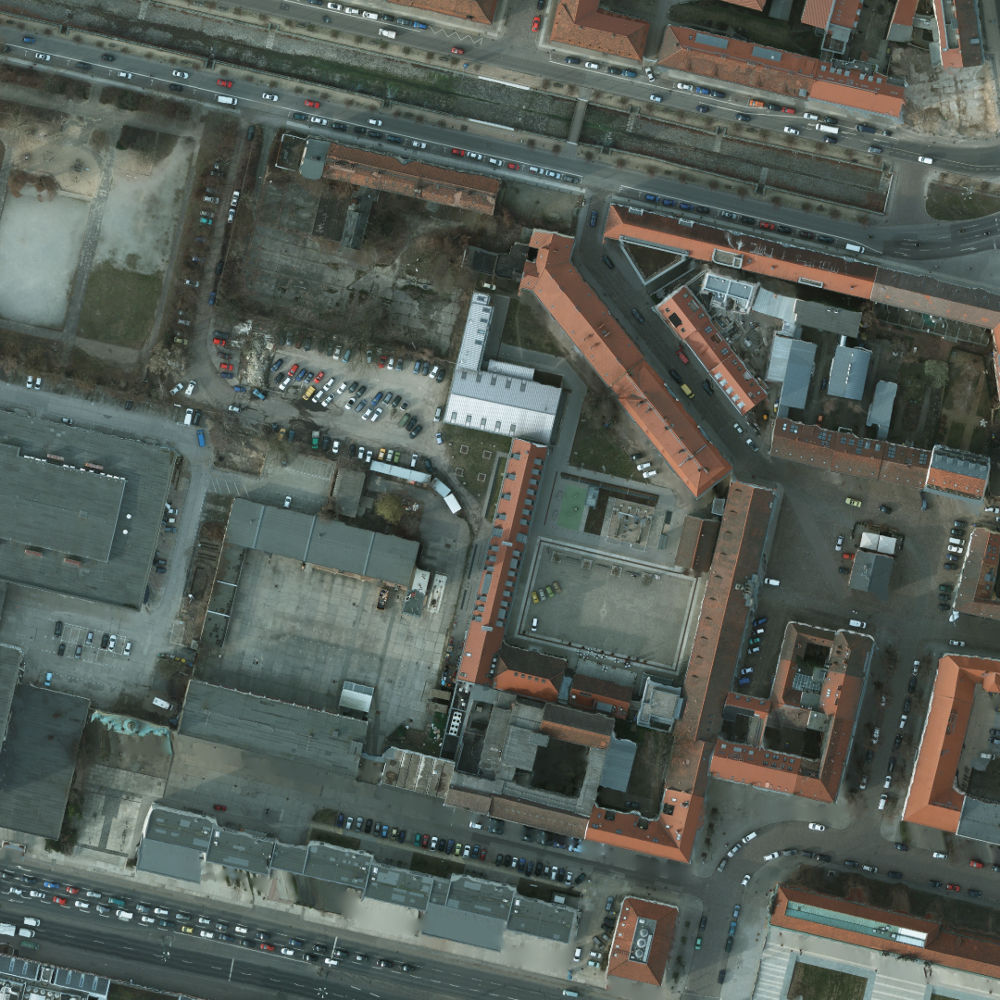}
    \caption{RGB data}
  \end{subfigure}
  \hspace{0.1\textwidth}
  \begin{subfigure}{0.35\textwidth}
    \includegraphics[width=\textwidth]{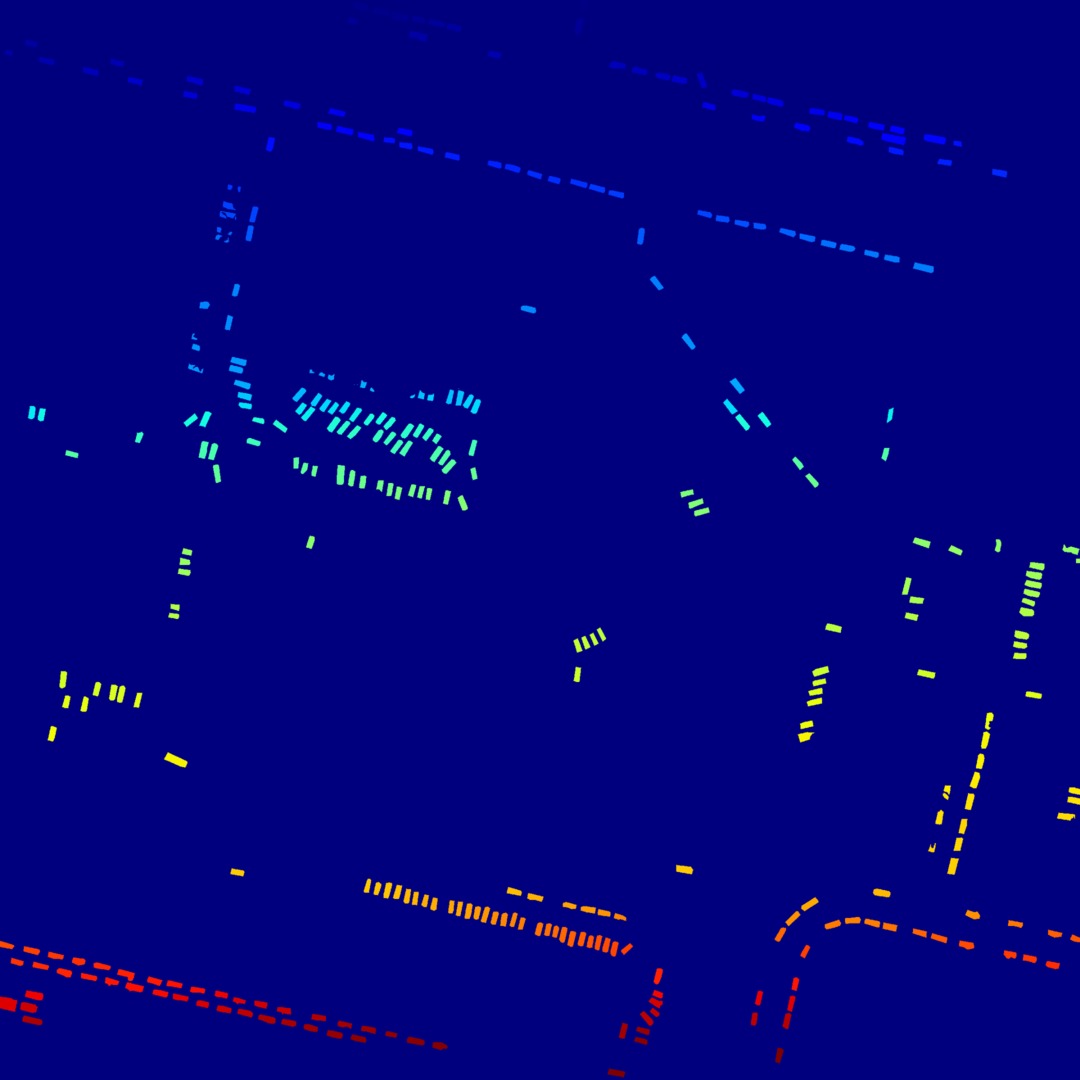}
    \caption{Vehicle ground truth}
  \end{subfigure}
  \begin{subfigure}{0.35\textwidth}
    \includegraphics[width=\textwidth]{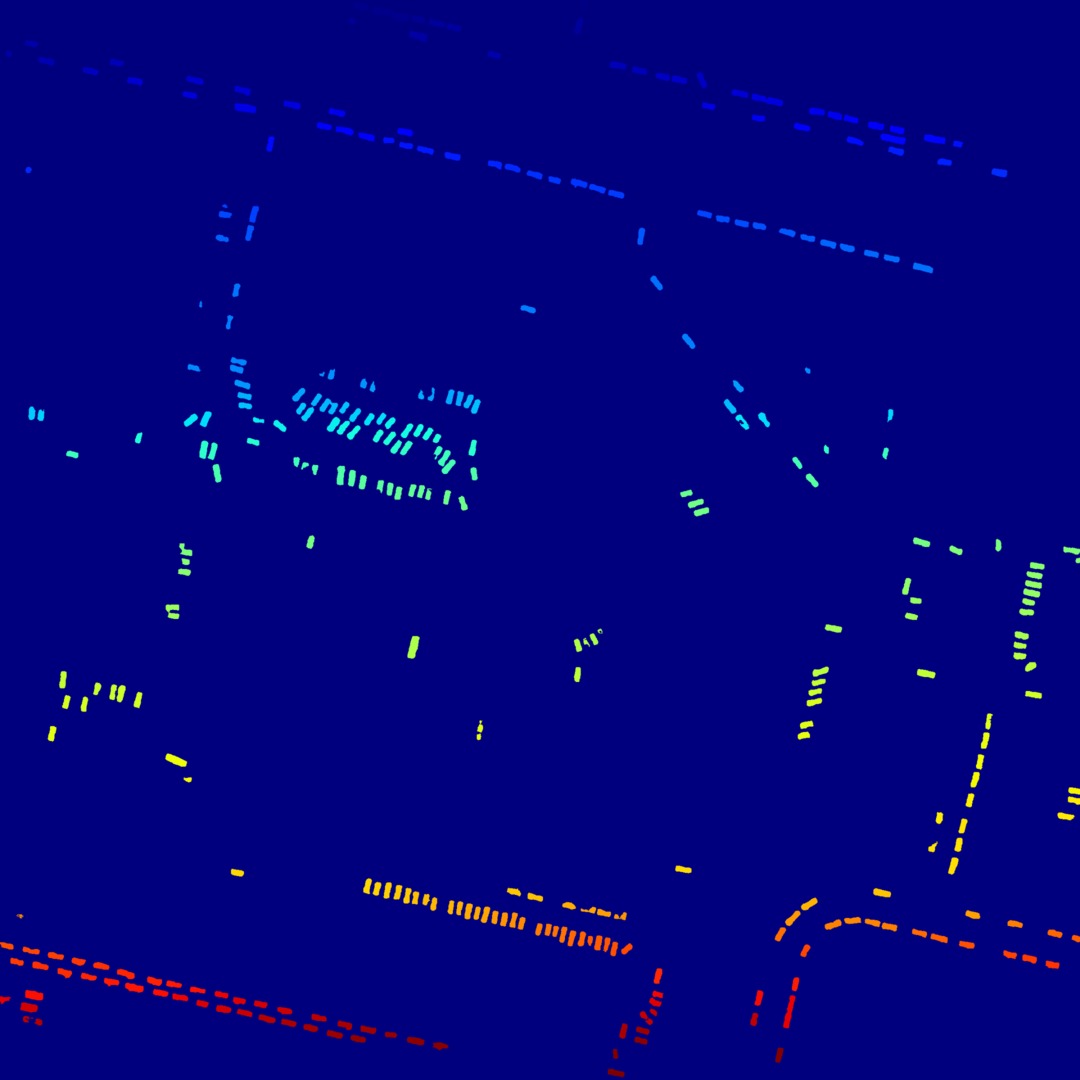}
    \caption{Predicted vehicles}
  \end{subfigure}
  \hspace{0.1\textwidth}
  \begin{subfigure}{0.35\textwidth}
    \includegraphics[width=\textwidth]{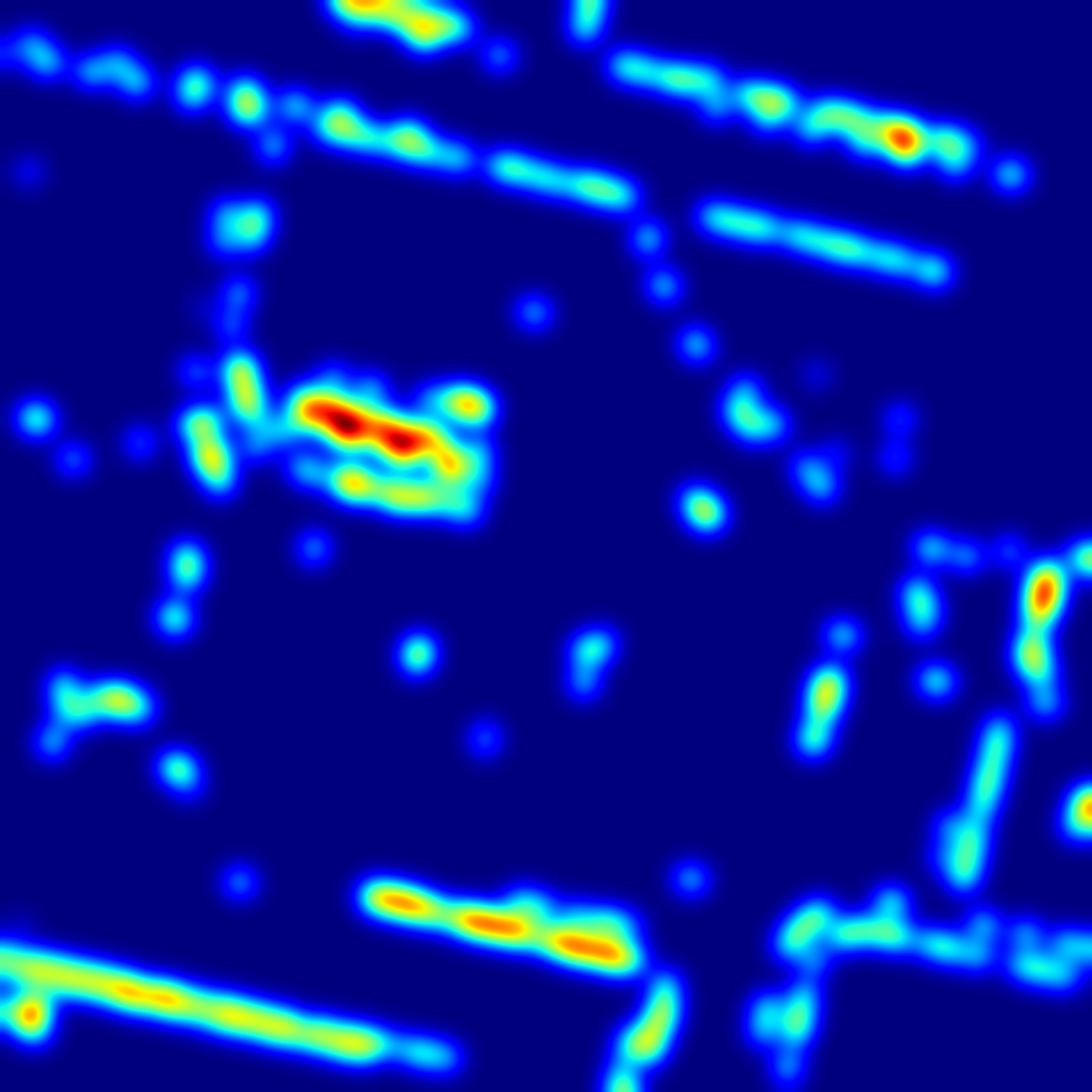}
    \caption{Vehicle occupancy heat map}
  \end{subfigure}
    \begin{subfigure}{0.35\textwidth}
    \includegraphics[width=\textwidth]{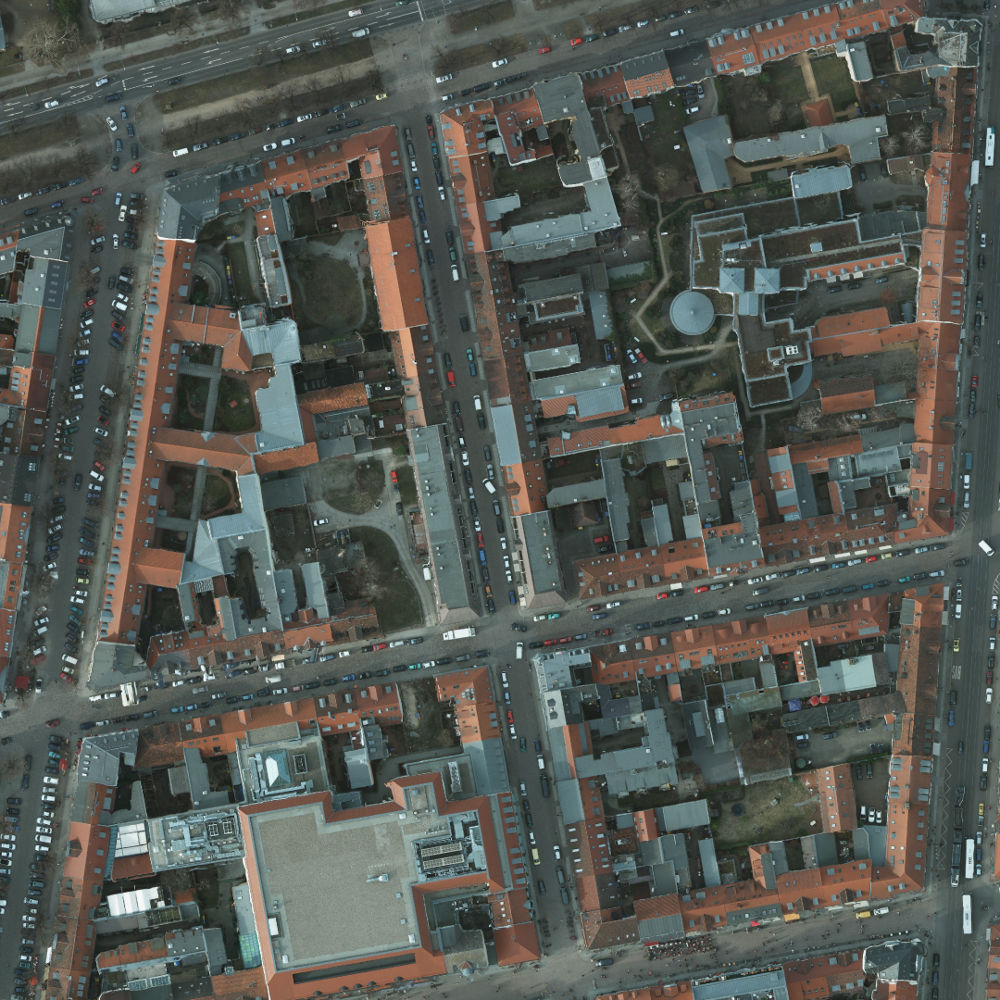}
    \caption{RGB data}
  \end{subfigure}
  \hspace{0.1\textwidth}
  \begin{subfigure}{0.35\textwidth}
    \includegraphics[width=\textwidth]{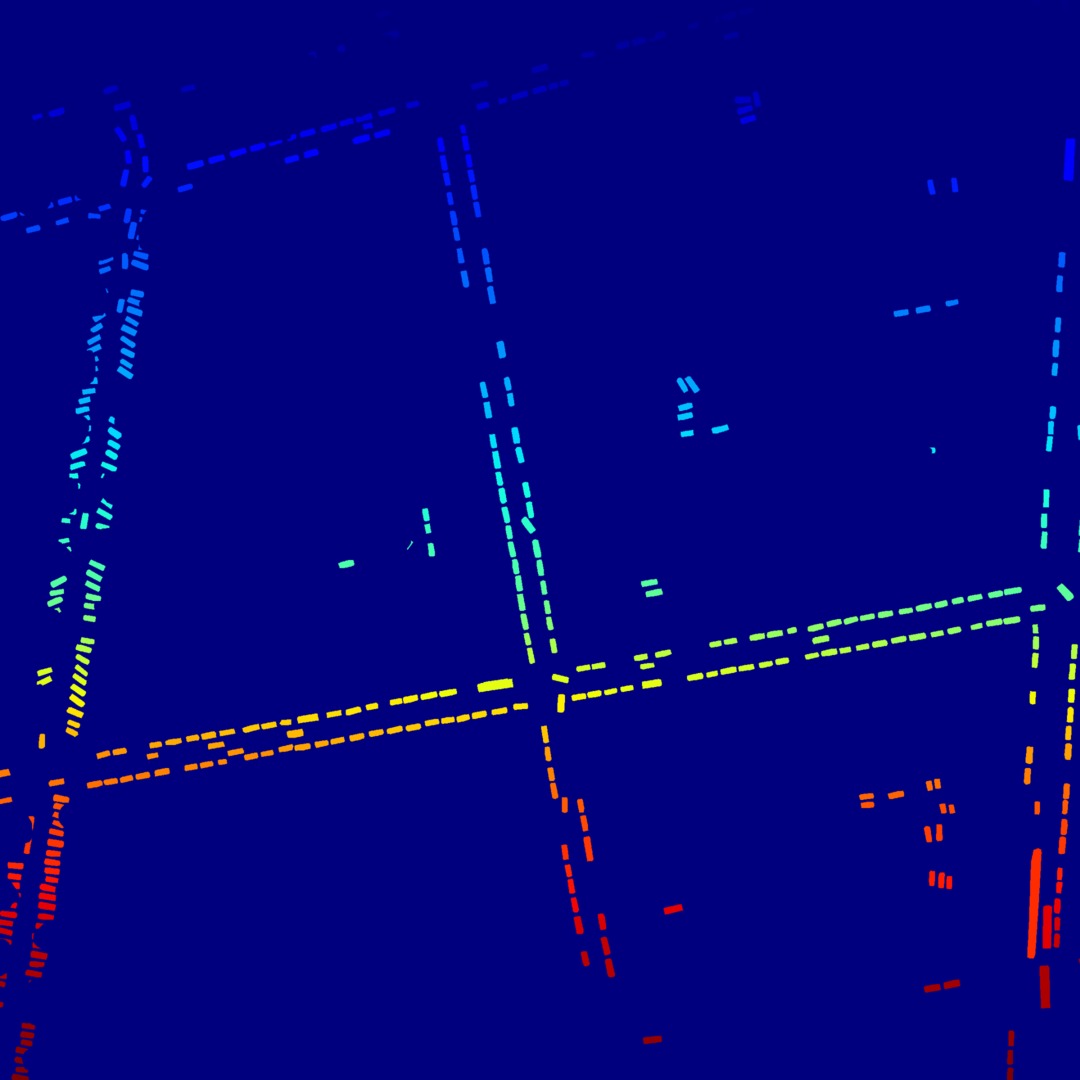}
    \caption{Vehicle ground truth}
  \end{subfigure}
  \begin{subfigure}{0.35\textwidth}
    \includegraphics[width=\textwidth]{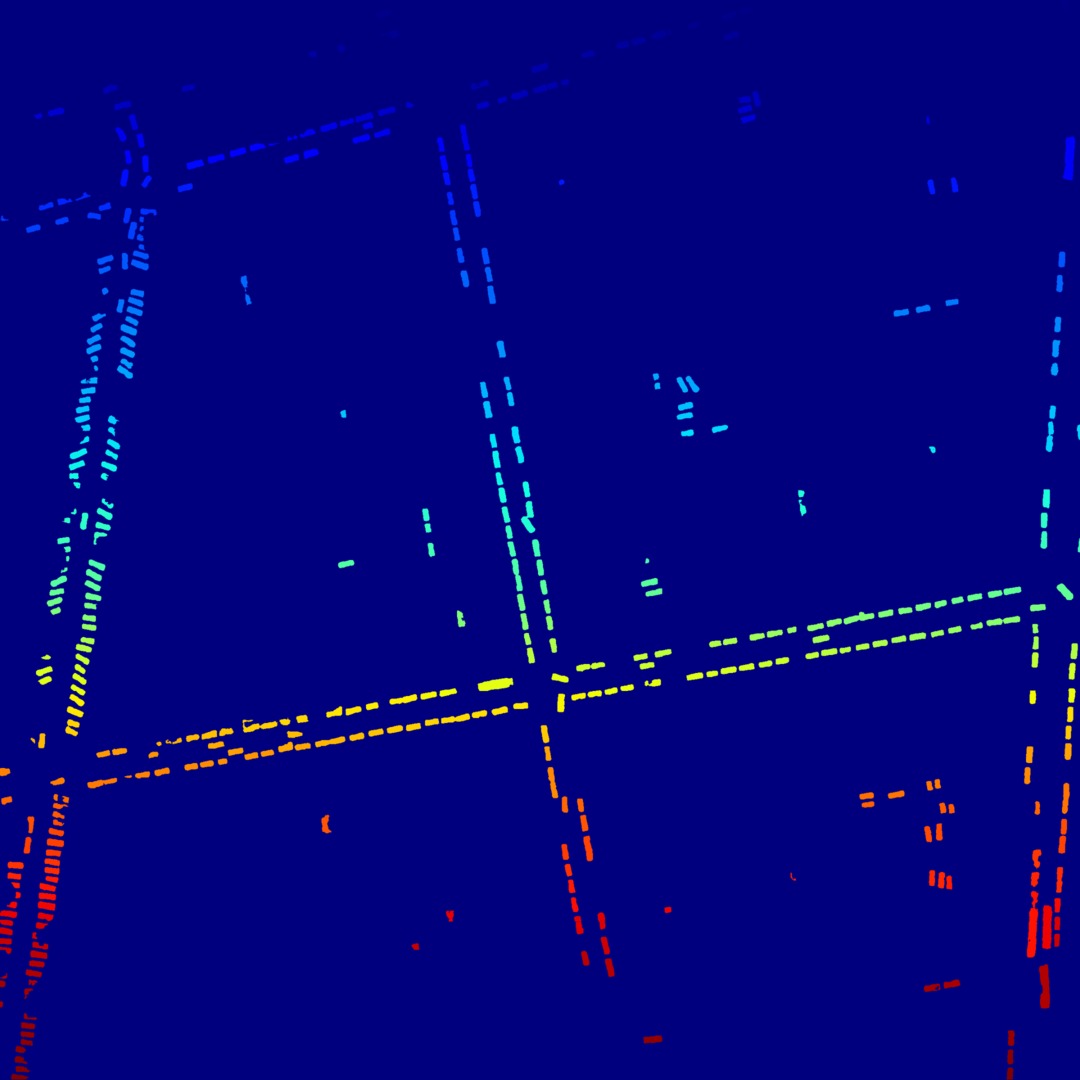}
    \caption{Predicted vehicles}
  \end{subfigure}
  \hspace{0.1\textwidth}
  \begin{subfigure}{0.35\textwidth}
    \includegraphics[width=\textwidth]{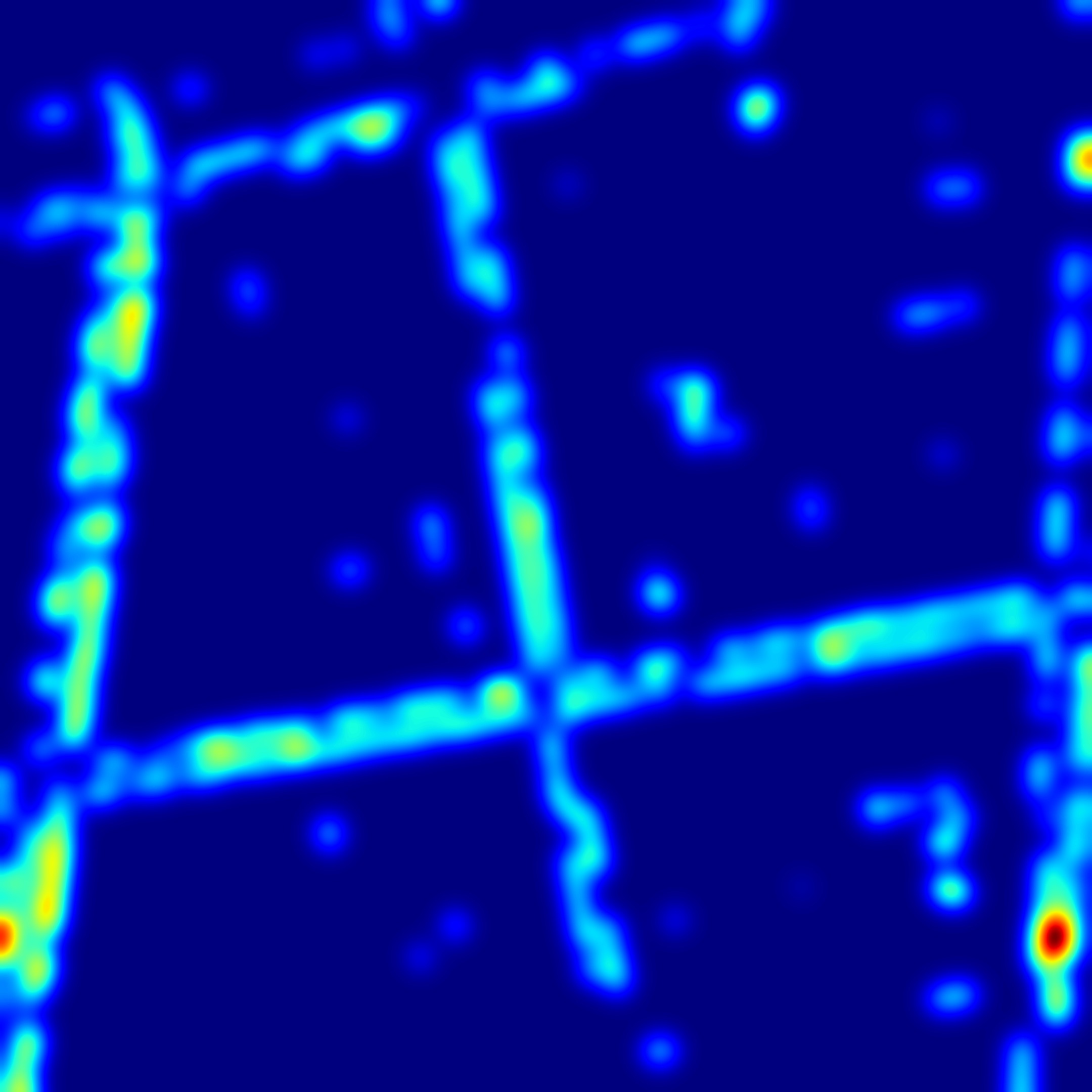}
    \caption{Vehicle occupancy heat map}
  \end{subfigure}
  \caption{Visualizing vehicles in the ISPRS Potsdam dataset (best viewed in color)}
  \label{fig:vehicle_visualization}
\end{figure*}

We also show that we can use this information to generate heat maps of the vehicle repartition in the tile (cf. \Cref{fig:vehicle_visualization}). Such a visualization provides an indicator of the vehicle density in the urban area, and exacerbates high traffic roads and ``hot spots'' corresponding to parking lots around points of interest (e.g. hospitals or shopping malls). This can be useful for many urban planning applications, such as parking lot sizing, traffic estimation, etc. In temporal data, this could be used to monitor traffic flows, but also study the impact of the road blocks, jams, etc.

\paragraph{Classification}
Using our AlexNet-based CNN classifier, we are able to classify the candidate vehicle instances generated by SegNet on the Potsdam dataset. We compare the predicted results to our enhanced vehicle ground truth. On our testing set, we classify correctly 75.0\% of the instances. We are able to improve even further this result by switching to the VGG-16 based classifier which brings accuracy to 77.3\% (cf. \Cref{table:vgg_classif} for detailed results). It should be noted that the dataset is predominantly comprised of cars, and that trucks and pick ups are less represented, which decreases their influence on the global accuracy score. \Cref{fig:positive_examples} illustrates some vehicle instances where our deep network-based segmentation and classification pipeline was successful, while \Cref{fig:negative_examples} shows some examples of correct segmentation but subsequent misclassification.

The fact that the average accuracy of the models on Potsdam are lower than results reported on VEDAI suggests that our networks suffered from overfitting. One hypothesis that we make is that VEDAI and Potsdam are images taken from two similar but subtly different environments. Indeed, Potsdam is a urban european city whereas VEDAI images have been shot over Utah, on a more rural american environment. Therefore, vehicle brands and different land covers around the cars might influence the classifiers. More comprehensive regularization (e.g. dropout \cite{Srivastava:2014:DSW:2627435.2670313}) during fine-tuning and/or training on a more diverse dataset would help alleviate this phenomenon.

\begin{table}
  \caption{Classification results on the enhanced Potsdam vehicle ground truth}
  \label{table:vgg_classif}
  \begin{tabularx}{0.5\textwidth}{c Y Y Y Y Y}
  \toprule
  Class & Car & Van & Truck & Pick up & Global\\
  \midrule
  AlexNet & 81\% & 42\% & \textbf{50\%} & \textbf{50\%} & 75\%\\
  VGG & \textbf{83\%} & \textbf{55\%} & \textbf{50\%} & 33\% & \textbf{77\%}\\
  \bottomrule
  \end{tabularx}
\end{table}

\section{CONCLUSION}\label{sec:conclusion}
In this work, we presented a two-step framework to detect, segment and classify vehicles from aerial RGB images using deep learning. More precisely, we showed that deep network designed for semantic segmentation such as SegNet are useful for scene understanding of remote sensing data and can be used to segment even small objects, such as cars and trucks. We reported results on the ISPRS Potsdam dataset showing very promising results on wheeled vehicles, achieving a F1 score of 77.3\% on this specific class.

In addition, we presented a simple deep learning based method to improve further this analysis by classifying the different types of vehicle present in the scene. We trained several deep CNN on the VEDAI dataset and transferred their knowledge to the Potsdam dataset. Our best model, fine-tuned from VGG-16, was able to classify successfully more than 77\% of the vehicles in our testing set.

Finally, this work meant to provide useful pointers for applying deep learning to scene understanding in Earth Observation with an object-oriented approach. We showed that not only deep networks are the state-of-the-art for semantic mapping, but that they can also be used to extract useful information at an object level, with a direct application on vehicle detection and classification.

We showed that it is possible to enumerate and extract individual object instances from the semantic map in order to analyze the vehicle distribution in the images, leading to the localization of points of interest such as high traffic roads and parking lots. This has many applications in traffic monitoring and urban planning, such as analyzing parking lots occupancy, finding pollutants vehicles in unauthorized zones (combined with a classifier), etc.

\section*{ACKNOWLEDGEMENTS}\label{ACKNOWLEDGEMENTS}

The Potsdam data set was provided by the German Society for Photogrammetry, Remote Sensing and Geoinformation (DGPF): \url{http://www.ifp.uni-stuttgart.de/dgpf/DKEP-Allg.html}.\\
The Vehicle Detection in Aerial Imagery (VEDAI) dataset was provided by S. Razakarivony and F. Jurie.\\
N. Audebert' work is funded by ONERA-TOTAL research project Naomi.\\
The authors would like to thank A. Boulch and A. Chan Hon Tong for fruitful discussions on object detection and classification.

{
	\begin{spacing}{0.9}
		\bibliography{GEOBIA16} 

\begin{thebibliography}{xx}

\bibitem[Audebert et al., 2016]{audebert_how_2016}
Audebert, N., Le~Saux, B. and Lefèvre, S., 2016.
\newblock How {Useful} is {Region}-based {Classification} of {Remote} {Sensing}
  {Images} in a {Deep} {Learning} {Framework}?
\newblock In: \emph{{IEEE} {International} {Geosciences} and {Remote} {Sensing}
  {Symposium} ({IGARSS})}, Beijing, China.

\bibitem[Badrinarayanan et al., 2015]{badrinarayanan_segnet:_2015}
Badrinarayanan, V., Kendall, A. and Cipolla, R., 2015.
\newblock {SegNet}: {A} {Deep} {Convolutional} {Encoder}-{Decoder}
  {Architecture} for {Image} {Segmentation}.
\newblock {\em arXiv preprint arXiv:1511.00561}.

\bibitem[Chen et al., 2014]{chen_vehicle_2014}
Chen, X., Xiang, S., Liu, C.~L. and Pan, C.~H., 2014.
\newblock Vehicle {Detection} in {Satellite} {Images} by {Hybrid} {Deep}
  {Convolutional} {Neural} {Networks}.
\newblock {\em IEEE Geoscience and Remote Sensing Letters} 11(10),
  pp.~1797--1801.

\bibitem[Cramer, 2010]{cramer_dgpf_2010}
Cramer, M., 2010.
\newblock The {DGPF} test on digital aerial camera evaluation – overview and
  test design.
\newblock {\em Photogrammetrie – Fernerkundung – Geoinformation} 2,
  pp.~73--82.

\bibitem[Everingham et al., 2014]{everingham_pascal_2014}
Everingham, M., Eslami, S. M.~A., Gool, L.~V., Williams, C. K.~I., Winn, J. and
  Zisserman, A., 2014.
\newblock The {Pascal} {Visual} {Object} {Classes} {Challenge}: {A}
  {Retrospective}.
\newblock {\em International Journal of Computer Vision} 111(1), pp.~98--136.

\bibitem[Ioffe and Szegedy, 2015]{ioffe_batch_2015}
Ioffe, S. and Szegedy, C., 2015.
\newblock Batch {Normalization}: {Accelerating} {Deep} {Network} {Training} by
  {Reducing} {Internal} {Covariate} {Shift}.
\newblock In: \emph{Proceedings of {The} 32nd {International} {Conference} on
  {Machine} {Learning}}, pp.~448--456.

\bibitem[Krizhevsky et al., 2012]{krizhevsky_imagenet_2012}
Krizhevsky, A., Sutskever, I. and Hinton, G.~E., 2012.
\newblock {ImageNet} {Classification} with {Deep} {Convolutional} {Neural}
  {Networks}.
\newblock In: F.~Pereira, C.~J.~C. Burges, L.~Bottou and K.~Q. Weinberger
  (eds), \emph{Advances in {Neural} {Information} {Processing} {Systems} 25},
  Curran Associates, Inc., pp.~1097--1105.

\bibitem[Lagrange et al., 2015]{lagrange_benchmarking_2015}
Lagrange, A., Le~Saux, B., Beaupere, A., Boulch, A., Chan-Hon-Tong, A., Herbin,
  S., Randrianarivo, H. and Ferecatu, M., 2015.
\newblock Benchmarking classification of earth-observation data: {From}
  learning explicit features to convolutional networks.
\newblock In: \emph{Geoscience and {Remote} {Sensing} {Symposium} ({IGARSS}),
  2015 {IEEE} {International}}, pp.~4173--4176.

\bibitem[Lecun et al., 1998]{lecun_gradient-based_1998}
Lecun, Y., Bottou, L., Bengio, Y. and Haffner, P., 1998.
\newblock Gradient-based learning applied to document recognition.
\newblock {\em Proceedings of the IEEE} 86(11), pp.~2278--2324.

\bibitem[Liang-Chieh et al., 2015]{liang-chieh_semantic_2015}
Liang-Chieh, C., Papandreou, G., Kokkinos, I., Murphy, K. and Yuille, A., 2015.
\newblock Semantic {Image} {Segmentation} with {Deep} {Convolutional} {Nets}
  and {Fully} {Connected} {CRFs}.
\newblock In: \emph{Proceedings of the {International} {Conference} on
  {Learning} {Representations}}.

\bibitem[Lin et al., 2014]{lin_microsoft_2014}
Lin, T.-Y., Maire, M., Belongie, S., Hays, J., Perona, P., Ramanan, D.,
  Dollár, P. and Zitnick, C.~L., 2014.
\newblock Microsoft {COCO}: {Common} {Objects} in {Context}.
\newblock In: D.~Fleet, T.~Pajdla, B.~Schiele and T.~Tuytelaars (eds),
  \emph{Computer {Vision} – {ECCV} 2014}, Lecture {Notes} in {Computer}
  {Science}, Springer International Publishing, pp.~740--755.
\newblock DOI: 10.1007/978-3-319-10602-1\_48.

\bibitem[Long et al., 2015]{long_fully_2015}
Long, J., Shelhamer, E. and Darrell, T., 2015.
\newblock Fully {Convolutional} {Networks} for {Semantic} {Segmentation}.
\newblock In: \emph{Proceedings of the {IEEE} {Conference} on {Computer}
  {Vision} and {Pattern} {Recognition}}, pp.~3431--3440.

\bibitem[Marmanis et al., 2016]{marmanis_semantic_2016}
Marmanis, D., Wegner, J.~D., Galliani, S., Schindler, K., Datcu, M. and Stilla,
  U., 2016.
\newblock Semantic {Segmentation} of {Aerial} {Images} with an {Ensemble} of
  {CNNs}.
\newblock {\em ISPRS Annals of Photogrammetry, Remote Sensing and Spatial
  Information Sciences} 3, pp.~473--480.

\bibitem[Nogueira et al., 2016]{nogueira_towards_2016}
Nogueira, K., Penatti, O. A.~B. and Dos~Santos, J.~A., 2016.
\newblock Towards {Better} {Exploiting} {Convolutional} {Neural} {Networks} for
  {Remote} {Sensing} {Scene} {Classification}.
\newblock {\em arXiv:1602.01517 [cs]}.
\newblock arXiv: 1602.01517.

\bibitem[Paisitkriangkrai et al., 2015]{paisitkriangkrai_effective_2015}
Paisitkriangkrai, S., Sherrah, J., Janney, P. and Hengel, A. V.-D., 2015.
\newblock Effective semantic pixel labelling with convolutional networks and
  {Conditional} {Random} {Fields}.
\newblock In: \emph{Proceedings of the {IEEE} {Conference} on {Computer}
  {Vision} and {Pattern} {Recognition} {Workshops}}, pp.~36--43.

\bibitem[Penatti et al., 2015]{penatti_deep_2015}
Penatti, O., Nogueira, K. and Dos~Santos, J., 2015.
\newblock Do deep features generalize from everyday objects to remote sensing
  and aerial scenes domains?
\newblock In: \emph{Proceedings of the {IEEE} {Conference} on {Computer}
  {Vision} and {Pattern} {Recognition} {Workshops}}, pp.~44--51.

\bibitem[Razakarivony and Jurie, 2016]{razakarivony_vehicle_2016}
Razakarivony, S. and Jurie, F., 2016.
\newblock Vehicle {Detection} in {Aerial} {Imagery}: {A} small target detection
  benchmark.
\newblock {\em Journal of Visual Communication and Image Representation} 34,
  pp.~187--203.

\bibitem[Rottensteiner et al., 2012]{rottensteiner_isprs_2012}
Rottensteiner, F., Sohn, G., Jung, J., Gerke, M., Baillard, C., Benitez, S. and
  Breitkopf, U., 2012.
\newblock The {ISPRS} benchmark on urban object classification and 3d building
  reconstruction.
\newblock {\em ISPRS Ann. Photogramm. Remote Sens. Spat. Inf. Sci} 1, pp.~3.

\bibitem[Russakovsky et al., 2015]{russakovsky_imagenet_2015}
Russakovsky, O., Deng, J., Su, H., Krause, J., Satheesh, S., Ma, S., Huang, Z.,
  Karpathy, A., Khosla, A., Bernstein, M., Berg, A.~C. and Fei-Fei, L., 2015.
\newblock {ImageNet} {Large} {Scale} {Visual} {Recognition} {Challenge}.
\newblock {\em International Journal of Computer Vision} 115(3), pp.~211--252.

\bibitem[Simonyan and Zisserman, 2014]{simonyan_very_2014}
Simonyan, K. and Zisserman, A., 2014.
\newblock Very {Deep} {Convolutional} {Networks} for {Large}-{Scale} {Image}
  {Recognition}.
\newblock {\em arXiv:1409.1556 [cs]}.
\newblock arXiv: 1409.1556.

\bibitem[Srivastava et al., 2014]{Srivastava:2014:DSW:2627435.2670313}
Srivastava, N., Hinton, G., Krizhevsky, A., Sutskever, I. and Salakhutdinov,
  R., 2014.
\newblock Dropout: A simple way to prevent neural networks from overfitting.
\newblock {\em J. Mach. Learn. Res.} 15(1), pp.~1929--1958.

\end{thebibliography}
	\end{spacing}
}

\end{document}